\documentclass[11pt,letterpaper,logo,twocolumn]{style}

\usepackage[numbers]{natbib}
\usepackage{graphicx}
\usepackage{booktabs}
\usepackage{amsmath,amsfonts,amssymb}
\usepackage{subcaption}
\usepackage{multirow}
\usepackage{colortbl}
\usepackage{listings}
\usepackage{xparse}
\usepackage{fontawesome5}
\usepackage{threeparttable}

\graphicspath{{./}{fig/}{figures/}{plot/}{pdf/}{table/}}
\usepackage{amsthm}
\usepackage{tcolorbox}
\tcbuselibrary{skins,breakable}
\tcbuselibrary{listingsutf8}
\usepackage{titletoc}
\usepackage{pifont}
\usepackage{mathtools}
\usepackage{bbm}
\usepackage{makecell}
\usepackage{adjustbox}

\title{PalmClaw: A Native On-Device Agent Framework for Mobile Phones}
\runningtitle{PalmClaw: A Native On-Device Agent Framework for Mobile Phones}
\PublicDate{2026-09-03}

\author{%
  {\Authfont
    \textbf{Hongru Cai}\textsuperscript{1} \quad
    \textbf{Yongqi Li}\textsuperscript{1}\advisor \quad
    \textbf{Ran Wei}\textsuperscript{2} \quad
    \textbf{Wenjie Li}\textsuperscript{1}
  }\\
  {\Affilfont
    \textsuperscript{1} The Hong Kong Polytechnic University \quad
    \textsuperscript{2} Hangzhou Diagens Biotechnology Co., Ltd. \\
    \texttt{\{henry.hongrucai, liyongqi0\}@gmail.com}
  }
}

\begin{document}

\begin{abstract}
Large Language Model (LLM) agents have moved beyond generating responses to executing multi-step tasks by calling tools, observing the results, and iteratively deciding the next action. Most agent systems run on desktops or servers, which support tool use and task automation. Mobile devices are also important agent environments because they are widely accessible and contain users' data, sensors, and daily-use applications. Existing mobile agents mainly operate smartphones through graphical user interface (GUI) actions such as tapping, swiping, and typing, which often form long, interface-dependent sequences, cannot directly access device capabilities, and make execution boundaries difficult to define. We present \textbf{PalmClaw}, an open-source agent framework that runs natively on mobile phones and manages the sessions, memory, skills, tools, and agent loop directly on the device. PalmClaw exposes device capabilities as device tools with explicit arguments, structured results, and clearly defined execution boundaries. This design enables agents to use mobile capabilities directly while keeping each action explicit and controlled. Experiments show an 11.5\% relative improvement in task success and a 94.9\% reduction in completion time over the strongest baseline, with lower setup burden and traces illustrating how execution boundaries are applied. 
\end{abstract}

\newcommand{\TitleLinks}{%
\centering
    \vspace{6pt}
    {\noindent\absfont\fontsize{11}{13}\selectfont
    \faGithub\ GitHub: \url{https://github.com/ModalityDance/PalmClaw}\par}%
}


\maketitle

\section{Introduction}
\label{sec:introduction}

Recent Large Language Model (LLM) agents have moved beyond generating responses to executing tasks in external environments~\citep{yao2023react,schick2023toolformer,li2026clawsbench}. Given a complex user instruction, the agent can decompose the task into multiple steps, call tools, observe the results, and iteratively decide what to do next until the task is completed~\citep{openclawGettingStarted,gao2025agentscope10developercentricframework}. This iterative interaction enables agents not only to offer advice, but more importantly, to autonomously execute tasks such as information seeking~\citep{zhu2026gisa}, file management~\citep{xie2024osworld,li2026clawsbench}, and workflow automation~\citep{xu2025theagentcompany}.

\begin{figure}[t]
\setlength{\abovecaptionskip}{-0.01cm}
\centering
\includegraphics[width=1\linewidth]{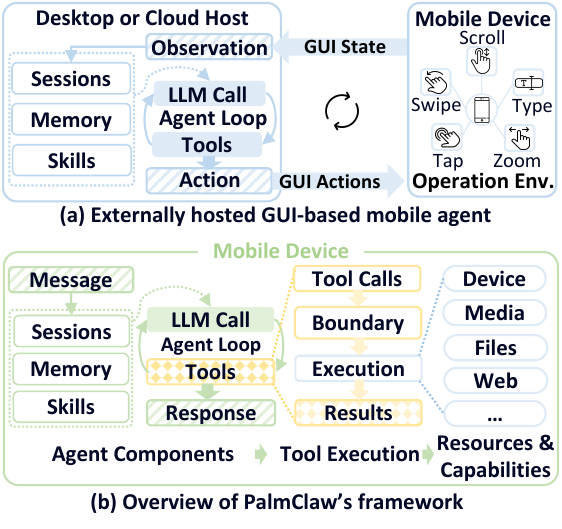}
\caption{Overview of PalmClaw. (a) An externally hosted mobile agent exchanges GUI states and 
GUI actions between a desktop or cloud host and the phone. (b) PalmClaw runs the agent components on the mobile device and connects them to device resources and capabilities through device tools.}
\vspace{-1.5em}
\label{fig:intro-overview}
\end{figure}

An agent is essentially a framework upon a foundational LLM through the incorporation of instructions, persistent memory, reusable skills, executable tools, session management, and an iterative agent loop. Together, these components allow the agent to retrieve relevant memory and session history, load task-specific instructions and skills, invoke the LLM, execute tools, and feed the resulting observations into subsequent turns. Representative agent frameworks, such as AutoGen~\citep{wu2023autogen}, AgentScope~\citep{gao2025agentscope10developercentricframework}, and OpenClaw~\citep{openclawGettingStarted} host these components on a desktop computer or cloud server.

While desktop and server platforms provide a rich and practical environment for agents, we argue that supporting agents natively on mobile devices is also important. First, mobile phones are among the most widely available personal computing devices, with four out of five people aged 10 or older worldwide owning one~\citep{itu2024mobileownership}. This broad availability allows mobile-native agents to reach users who may not have continuous access to a personal computer. Second, smartphones store highly personal data, provide sensors such as cameras, microphones, and location, and host many of the applications used in everyday life~\citep{wang2024mobileagentbenchefficientuserfriendlybenchmark}. Running agents directly on smartphones can therefore place them closer to the users, their device resources, and the capabilities required for daily tasks.

Despite this potential, existing mobile agents primarily operate smartphones through GUI actions such as tapping, swiping, and typing (see Figure~\ref{fig:intro-overview}(a))~\citep{wang2024mobileagentbenchefficientuserfriendlybenchmark,zhang-etal-2025-agentcpm,ye2025mobileagentv3fundamentalagentsgui}. Although broadly compatible with existing applications, this approach has three limitations: 1) high-level tasks become long action sequences that are sensitive to layout changes and complex interfaces; 2) it cannot directly access device resources and capabilities, such as local files and sensor data; and 3) GUI control grants broad access to reachable interfaces, without clear task-specific access limits or execution boundaries. GUI interaction is therefore useful for general screen operations, but insufficient as the sole action space for mobile agents. The recent rise of terminal-native agents such as Codex CLI~\citep{openaiCodexCLI}, Claude Code~\citep{anthropicClaudeCode}, and Gemini CLI~\citep{googleGeminiCLI} reflects a broader shift toward explicit and executable tool-based interaction, motivating a similar native on-device design for mobile agents.

We present \textbf{PalmClaw}, an open-source agent framework designed to run natively within the mobile environment\footnote{PalmClaw is released under the AGPLv3 license. Installable packages are available at \url{https://github.com/ModalityDance/PalmClaw/releases/latest}}. As shown in Figure~\ref{fig:intro-overview}(b), PalmClaw runs the agent loop and manages its memory, skills, tools, and session state \textbf{on the mobile device}. It exposes device resources and capabilities through \textbf{device tools}: device operations with explicit arguments, structured results, and tool-specific execution boundaries. Together, our framework reduces dependence on external desktop or server devices for agent orchestration, allows agents to use mobile capabilities more directly, and keeps each mobile action explicit and bounded. Experiments on representative mobile tasks show that PalmClaw improves task completion success by 11.5\% relative to the strongest baseline and reduces average task completion time by 94.9\%. Deployment and trace analyses further show lower setup burden and illustrate how execution boundaries are applied.

The key contributions are as follows:
\begin{list}{$\bullet$}{
\setlength{\leftmargin}{1em}
\setlength{\itemindent}{0pt}
\setlength{\itemsep}{0pt}
\setlength{\topsep}{0pt}
}
\item We present PalmClaw, an open-source mobile-native agent framework that hosts the agent loop, memory, skills, tools, and session state directly on the mobile device.

\item We design device tools that expose mobile resources and capabilities through explicit arguments, structured results, and tool-specific execution boundaries.

\item We evaluate PalmClaw on mobile tasks, showing an 11.5\% relative improvement in task success and a 94.9\% reduction in completion time, with lower setup burden and a trace-based analysis of execution boundaries.

\end{list}

\section{Related Work}
\label{sec:related-work}

\paragraph{General agent frameworks.}
LLM agents extend language models with tool use, intermediate observations, and multi-step action loops. ReAct~\citep{yao2023react} connects reasoning traces with actions, and Toolformer~\citep{schick2023toolformer} studies API calls as part of language modeling. Frameworks such as AutoGen~\citep{wu2023autogen}, AgentScope~\citep{gao2025agentscope10developercentricframework}, and OpenClaw~\citep{openclawGettingStarted} package these mechanisms into systems for real tasks, hosted on computers or cloud servers. PalmClaw instead hosts the agent components directly on the mobile device.

\begin{figure*}[!t]
\centering
\includegraphics[width=1\linewidth]{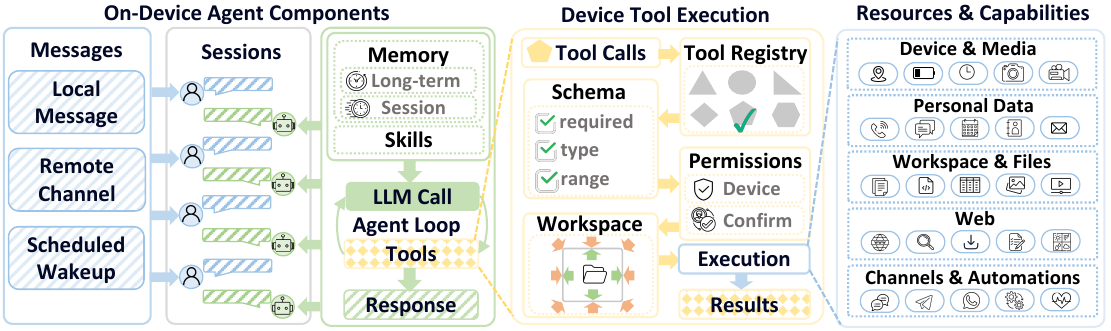}
\caption{Overview of the PalmClaw framework. Inputs are assigned to mobile sessions, where context is assembled, and the agent loop is executed. Tool calls are checked and executed on the device before their results are returned to the agent loop.}
\label{fig:framework-overview}
\end{figure*}

\paragraph{Mobile agents.}
Mobile agents study how LLM-based systems can complete tasks on smartphones. Existing work has developed mobile agents and benchmarks, including AndroidWorld~\citep{rawles2024androidworlddynamicbenchmarkingenvironment}, MobileAgentBench~\citep{wang2024mobileagentbenchefficientuserfriendlybenchmark}, Mobile-Bench~\citep{deng-etal-2024-mobile}, AgentCPM-GUI~\citep{zhang-etal-2025-agentcpm}, and Mobile-Agent-v3.5~\citep{xu2026mobile}. Related systems such as ApkClaw~\citep{apkclawSoftware}, MobileClaw~\citep{mobileclawSoftware}, and ClawMobile~\citep{du2026clawmobile} also explore phone-side task execution with GUI actions. PalmClaw focuses on hosting the agent framework on the phone and exposing device capabilities through device tools.


\section{Framework}
\label{sec:framework}

This section describes the agent components in PalmClaw and how they work together during on-device execution.

\subsection{Overview}
\label{sec:framework-overview}

PalmClaw runs the components needed for multi-step agent execution directly on the mobile device. As shown in Figure~\ref{fig:framework-overview}, sessions organize user interactions, memory preserves reusable information, skills provide task-specific instructions, and tools expose device resources and capabilities. The agent loop combines these components through repeated model and tool interactions. LLM inference is provided through a remote LLM API, while context management, tool execution, and session state remain on the mobile device. PalmClaw integrates these components into a mobile application, with representative interfaces shown in Figure~\ref{fig:app-screenshots}.

\subsection{Agent Components}
\label{sec:framework-components}

PalmClaw maintains sessions, memory, skills, and device tools directly on the mobile device.

\paragraph{Sessions.}
Sessions organize persistent agent interactions from three input sources: local chat messages, messages from connected remote channels, and automated messages generated by scheduled wakeups. Each session stores its conversation history, tool traces, attachments, and a dedicated workspace for documents, temporary files, and generated artifacts. PalmClaw routes each input to an existing or new session, allowing tasks to continue across multiple turns while keeping concurrent tasks separate.

\paragraph{Memory.}
PalmClaw maintains two persistent memory layers: 1) shared long-term memory, which stores information available across sessions and is included in each agent's turn, and 2) per-session summaries, which keep compact records of task progress and conclusions. When unconsolidated messages reach a configurable window, a separate LLM call summarizes older messages and updates both layers. Memory tools allow the agent to read or update long-term memory and to read or search the history of a selected session.

\begin{figure}[t]
\centering
\includegraphics[width=1\linewidth]{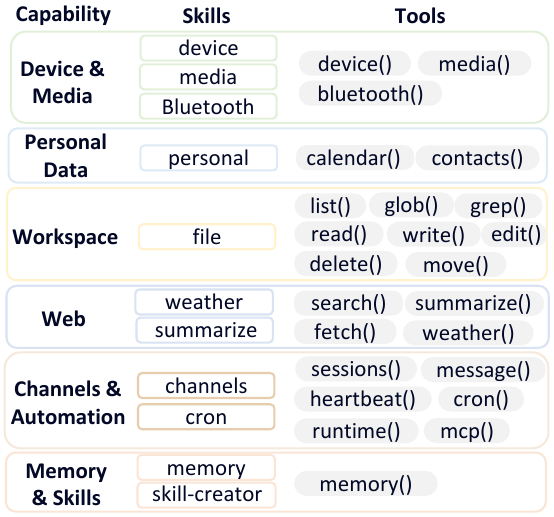}
\caption{Built-in skills and tool groups in PalmClaw. Skills provide reusable task instructions, while tools expose device, personal data, workspace, web, communication, and automation capabilities.}
\vspace{-1em}
\label{fig:skills-tools-map}
\end{figure}

\paragraph{Skills.}
Skills are reusable task instructions for applying tools and completing common routines, stored in \texttt{SKILL.md} files. PalmClaw provides built-in and user-installed skills. For each turn, it adds a summary of available skills to the context, loads the full instructions of skills marked as always active, and selects other relevant skills by matching recent user messages against their names, descriptions, and keywords. Only always-active and selected skills are loaded in full, providing task-specific guidance without loading every skill into the context. Figure~\ref{fig:skills-tools-map} summarizes the built-in skills and tool groups.

\paragraph{Tools.}
Device tools connect model-generated tool calls to resources and capabilities available on the phone. Each tool links a model-facing specification containing its name, description, and JSON argument schema and a device-facing function that invokes a mobile service or app-managed operation. PalmClaw sends the specification to the LLM, executes the associated function when called, and returns a structured result or error. Built-in device tools cover device status and permissions, media, Bluetooth, calendar, and contacts, and workspace files; other tool groups support web access, memory, communication, and scheduled automation.

\begin{figure}[!t]
\centering
\includegraphics[width=1\linewidth]{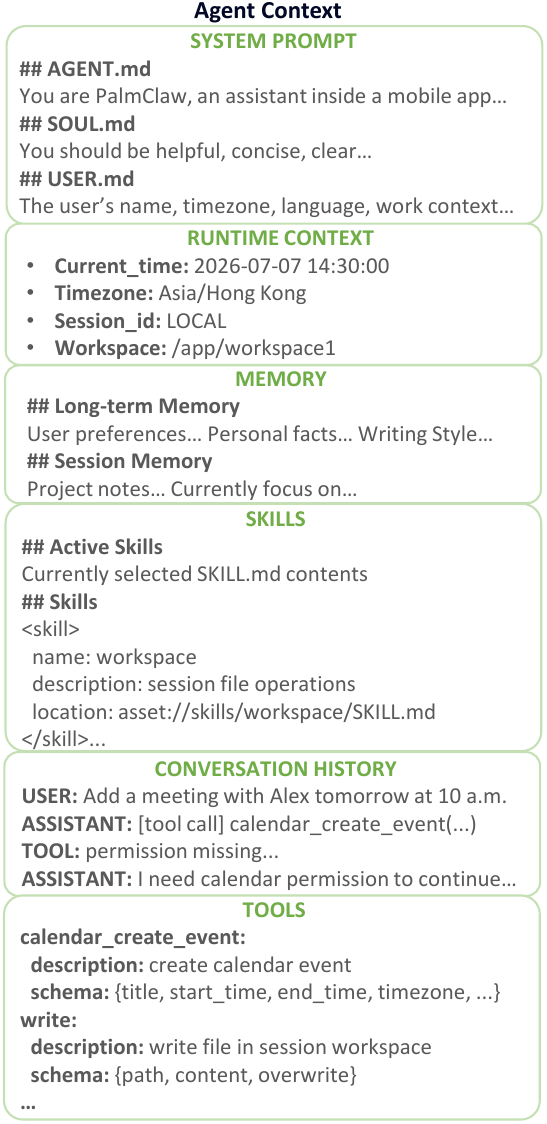}
\caption{Agent context for one PalmClaw turn. Model messages combine system instructions, runtime context, memory, skills, and conversation history; tool specifications are passed alongside them for tool calling.}
\label{fig:agent-context}
\label{fig:prompt-composition}
\end{figure}

\subsection{Agent Execution}
\label{sec:framework-execution}

These components work together through context assembly, the agent loop, and device tool execution.

\paragraph{Context assembly.}
After an input is assigned to a session, PalmClaw assembles the agent context from six sources: 1) system instructions from \texttt{AGENT.md}, \texttt{SOUL.md}, and \texttt{USER.md}; 2) runtime context, including the current time, timezone, session ID, and workspace paths; 3) shared long-term memory; 4) active skill instructions and a summary of available skills; 5) recent conversation history, attachments, and tool traces; and 6) available tool specifications, including their names, descriptions, and argument schemas. The first five sources form the model messages, while the tool specifications are provided alongside them to the remote LLM API. Figure~\ref{fig:agent-context} shows an example of this context.

\paragraph{Agent loop.}
PalmClaw runs each agent turn as a bounded multi-round loop that connects remote LLM reasoning with on-device tool execution~\citep{yao2023react,schick2023toolformer}. In each round, PalmClaw rebuilds the context from the current session state, sends the model messages and available tool specifications to the remote LLM API, and stores the returned assistant message in the session. If the response contains tool calls, PalmClaw executes them, appends their results to the session, and starts the next round with the updated context. If the response contains no tool calls, it becomes the final reply. The loop also ends when a tool signals task completion or the round limit is reached.

\paragraph{Tool execution.}
Tool calls from the agent loop enter the device tool execution path shown in Figure~\ref{fig:framework-overview}. PalmClaw first matches each call to the tool registry. Before the corresponding operation is executed, the call passes three checks: 1) \textit{Schema} validates required arguments, types, and allowed ranges. 2) \textit{Permissions} includes device authorization and user confirmation. Missing permissions invoke the system permission or settings flow, while actions requiring direct user involvement proceed only after the user confirms or completes them. 3) \textit{Workspace} resolves file paths within the current session workspace or approved shared storage, rejects other session workspaces and unapproved global paths, and requires confirmation before external writes. A call that passes these checks is executed against the corresponding device resource or capability. Its structured result is returned to the session for the next round; if a check fails or the user declines, PalmClaw returns a structured error.

\section{Evaluation}
\label{sec:evaluation}

We evaluate PalmClaw's on-device agent framework through three questions:
\begin{list}{$\bullet$}{
    \setlength{\leftmargin}{1em}
    \setlength{\itemindent}{0pt}
    \setlength{\itemsep}{0pt}
    \setlength{\topsep}{0pt}
}
    \item \textbf{RQ1:} Does PalmClaw support effective mobile task completion through its on-device agent framework and device tools?
    \item \textbf{RQ2:} Does hosting the agent framework on the mobile device reduce external deployment and operation requirements?
    \item \textbf{RQ3:} How are execution boundaries applied when PalmClaw's device tools execute mobile actions?
\end{list}

\subsection{Experimental Setup}
\label{sec:eval-setup}

\paragraph{Datasets.}
We evaluate PalmClaw on two datasets. 1) \textbf{MobileTask} contains 70 mobile tasks adapted from AndroidWorld~\citep{rawles2024androidworlddynamicbenchmarkingenvironment}, MobileAgentBench~\citep{wang2024mobileagentbenchefficientuserfriendlybenchmark}, and Mobile-Bench~\citep{deng-etal-2024-mobile}. It covers calendar, weather, contacts, audio, notes, files, Bluetooth, and media tasks, and is evaluated by task success. We use deterministic final-state checks when possible, evidence-grounded judging for answer-based tasks, and manual review for permission- or device-state tasks. 2) \textbf{AssistantBench}~\citep{yoran2024assistantbenchwebagentssolve} evaluates realistic web information-seeking tasks with verifiable answers. We use a 19-task development-set subset because the local test split does not provide gold answers, and some development tasks contain unstable real-time values. AssistantBench is evaluated with the official accuracy metric. Table~\ref{tab:dataset-stats} summarizes the dataset composition, and Appendix~\ref{app:datasets} gives construction and metric details.

\begin{table}[t]
\caption{Dataset statistics. MobileTask lists task types and source benchmarks; AssistantBench lists difficulty levels from the dev set.}
\label{tab:dataset-stats}
\centering
\small
\setlength{\tabcolsep}{2pt}
\begin{tabular*}{\linewidth}{@{}p{0.28\linewidth}p{0.18\linewidth}p{0.34\linewidth}@{\extracolsep{\fill}}r@{}}
\toprule
\textbf{Dataset} & \textbf{Category} & \textbf{Source} & \textbf{\#} \\
\midrule
\multirow{12}{*}{MobileTask} & \multirow{2}{*}{Calendar} & AndroidWorld & 32 \\
 &  & MobileAgentBench & 3 \\
\cmidrule(lr){2-4}
 & Weather & Mobile-Bench & 10 \\
\cmidrule(lr){2-4}
 & \multirow{2}{*}{Contacts} & AndroidWorld & 3 \\
 &  & MobileAgentBench & 4 \\
\cmidrule(lr){2-4}
 & \multirow{2}{*}{Audio} & AndroidWorld & 4 \\
 &  & MobileAgentBench & 3 \\
\cmidrule(lr){2-4}
 & Notes & MobileAgentBench & 4 \\
\cmidrule(lr){2-4}
 & Files & MobileAgentBench & 3 \\
\cmidrule(lr){2-4}
 & Bluetooth & AndroidWorld & 2 \\
\cmidrule(lr){2-4}
 & Media & AndroidWorld & 2 \\
\cmidrule(lr){2-4}
 & \textbf{Total} &  & \textbf{70} \\
\midrule
\multirow{3}{*}{AssistantBench} & Medium & Dev set & 10 \\
 & Hard & Dev set & 9 \\
\cmidrule(lr){2-4}
 & \textbf{Total} &  & \textbf{19} \\
\bottomrule
\end{tabular*}
\end{table}

\begin{table*}[t]
\caption{Task-completion results. MobileTask reports success rate and average actions, total tokens, and completion time per task; AssistantBench reports official accuracy on the selected dev subset.}
\label{tab:main-results}
\centering
\begingroup
\small
\setlength{\tabcolsep}{2.5pt}
\begin{tabular*}{\linewidth}{
    @{\extracolsep{\fill}}
    l
    c
    c
    ccccc
    @{}
}
\toprule
\multirow{2}{*}{\textbf{Framework}} & \multirow{2}{*}{\textbf{Host}} & \multirow{2}{*}{\textbf{Action space}} & \multicolumn{4}{c}{\textbf{MobileTask}} & \textbf{AssistantBench} \\
\cmidrule(lr){4-7}\cmidrule(lr){8-8}
 & & & \textbf{Actions $\downarrow$} & \textbf{Tokens $\downarrow$} & \textbf{Time $\downarrow$} & \textbf{Success Rate $\uparrow$} & \textbf{Accuracy $\uparrow$} \\
\midrule
MobileClaw & Desktop & GUI actions & 94.3 & 1.30M & 1197.0s & 60.0\% & 5.26\% \\
ClawMobile & Termux & APIs and GUI actions & 51.4 & 466.4K & 451.1s & 77.1\% & 10.49\% \\
ApkClaw & Mobile app & GUI actions & 103.9 & 2.06M & 348.8s & 87.1\% & 25.79\% \\
PalmClaw & Mobile app & Device tools & \textbf{2.8} & \textbf{50.4K} & \textbf{17.7s} & \textbf{97.1\%} & \textbf{36.85\%} \\
\bottomrule
\end{tabular*}
\endgroup
\vspace{-1em}
\end{table*}

\paragraph{Baselines.}
We compare PalmClaw with three representative open-source mobile-agent baselines. 1) \textbf{ApkClaw}~\citep{apkclawSoftware} is an Android app for natural-language device control and acts through GUI operations such as tapping, swiping, typing, and opening apps. 2) \textbf{MobileClaw}~\citep{mobileclawSoftware} uses a phone-side client with an external Python runtime and controls the phone through screen observations and GUI actions. 3) \textbf{ClawMobile}~\citep{du2026clawmobile} places the agent runtime in Termux, an Android terminal environment that provides a Linux-like space on the phone, and executes actions through phone APIs, shell commands, and GUI control.

\paragraph{Implementation.}
All mobile experiments are run on Android, using a Xiaomi Redmi 2312DRAABC phone with Android 13 (SDK 33). We use \texttt{DeepSeek-V4-Flash}~\citep{deepseekai2026deepseekv4highlyefficientmilliontoken} as the LLM backbone for all systems. MobileClaw additionally requires a visual GUI model for screenshot-based grounding, so we use \texttt{qwen3.7-plus}~\citep{qwen37}. For AssistantBench, PalmClaw uses a 30-round limit and screen-based baselines use a 100-round limit, since GUI agents often spend extra rounds on observation and navigation before taking an action.

\subsection{Task Completion (RQ1)}
\label{sec:eval-task-completion}

We first test whether PalmClaw can complete mobile tasks through its on-device agent framework and device tools. In addition to task success, we report the average number of actions, total tokens, and completion time per task. We count one LLM API call as one action. Token counts include all input and output tokens, and completion time is averaged over all tasks. These measures are essential because users are sensitive to response latency when completing tasks on a phone.

Results in Table~\ref{tab:main-results} give two findings. 1) PalmClaw differs from the baselines in both framework placement and action space. It runs the agent components on the mobile device and acts through device tools, whereas the baselines rely on GUI actions, combine APIs with GUI control, or host the agent framework on a desktop. 2) On MobileTask, PalmClaw achieves an 11.5\% relative improvement in task success and a 94.9\% reduction in average completion time over the strongest baseline, while using fewer actions and tokens. It also obtains the highest accuracy on AssistantBench. These results reflect PalmClaw's complete system design, including its on-device framework placement and device-tool action space. The efficiency gains mainly come from device tools, which expose mobile capabilities as structured operations and avoid long sequences of screen observation, navigation, and GUI actions.

\subsection{Deployment and Operation (RQ2)}
\label{sec:eval-deployment}

\begin{table}[t]
\caption{Deployment and operation comparison. Computer indicates whether a separate desktop or server is required; CLI indicates command-line setup or operation; Bridge indicates a connection between the phone and another control service.}
\label{tab:deployment-operation}
\centering
\begingroup
\small
\setlength{\tabcolsep}{3pt}
\begin{tabular*}{\linewidth}{@{\extracolsep{\fill}}lccccc@{}}
\toprule
\multirow{2}{*}{\textbf{Framework}} & \multicolumn{3}{c}{\textbf{External Requirements}} & \multicolumn{2}{c}{\textbf{Setup}} \\
\cmidrule(lr){2-4}\cmidrule(lr){5-6}
 & \textbf{Computer} & \textbf{CLI} & \textbf{Bridge} & \textbf{\# Steps} & \textbf{Time} \\
\midrule
OpenClaw & Yes & Yes & Optional & 4 & $\sim$5 min \\
MobileClaw & Yes & Yes & Yes & 8 & $\sim$15 min \\
ClawMobile & No & Yes & Optional & 6 & $\sim$25 min \\
ApkClaw & No & No & No & 3 & $\sim$5 min \\
PalmClaw & No & No & No & 2 & $\sim$2 min \\
\bottomrule
\end{tabular*}
\endgroup
\end{table}

To examine whether PalmClaw reduces the external environment needed to deploy and operate a mobile agent, we compare two aspects of each framework: external requirements and setup burden. The comparison includes the mobile-agent baselines used above and OpenClaw~\citep{openclawGettingStarted}, which serves as a representative desktop-hosted agent framework. \textbf{Computer} records whether deployment or operation requires a separate computer. \textbf{CLI} denotes setup or operation through command-line commands. \textbf{Bridge} denotes a required connection between the phone and another runtime or control service. We then manually deploy each framework and record the visible setup steps and time.

Table~\ref{tab:deployment-operation} gives two findings. 1) PalmClaw does not require a separate computer, CLI workflow, or bridge. The agent can therefore be started and used on the mobile device without attaching the task flow to another execution environment. This follows from PalmClaw's on-device design: its agent components and execution flow are integrated within the same mobile application. 2) PalmClaw needs two setup steps and about two minutes before the first instruction, less than the compared frameworks. Because both the user entry point and agent framework are on the phone, setup does not require a separate computer or bridge.

\subsection{Execution Boundaries (RQ3)}
\label{sec:eval-boundary}

We use trace-based case analysis to illustrate how execution boundaries are applied during device tool execution. We collect records where PalmClaw stops, requests user involvement, or returns a failure because a tool call reaches a permission or confirmation boundary, a workspace boundary, or an unsupported operation.

\begin{figure}[!t]
\centering
\includegraphics[width=1\linewidth]{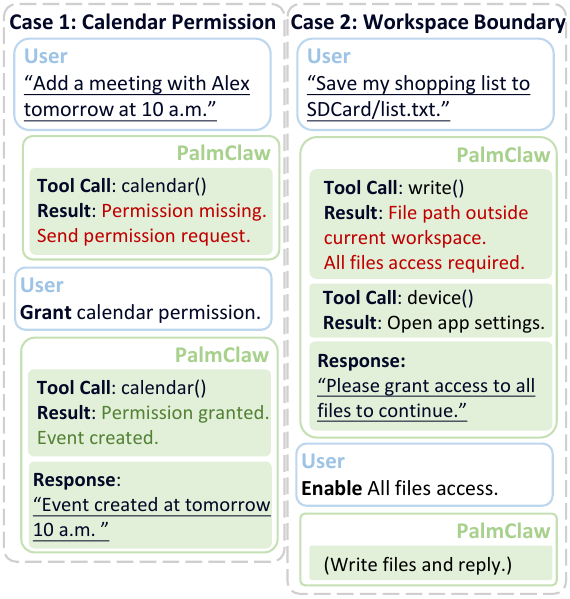}
\caption{Two execution-boundary traces. PalmClaw requests device permissions when possible and gives explicit settings instructions when a workspace boundary requires all files access.}
\label{fig:boundary-traces}
\end{figure}

The traces show how PalmClaw checks a tool call before it becomes a mobile operation. Each call is matched to the tool registry and then passes the relevant schema, permission or confirmation, and workspace checks before execution. Thus, the agent does not receive an open command channel to the phone. Figure~\ref{fig:boundary-traces} visualizes two representative cases: a calendar permission that can be granted through the platform permission flow, and a workspace boundary where a public path requires manual settings access. Table~\ref{tab:boundary-analysis} summarizes these cases and two additional boundary situations.

\begin{table*}[!t]
\caption{Representative execution-boundary cases. Each case shows the requested action, the boundary reached before execution, and PalmClaw's response.}
\label{tab:boundary-analysis}
\centering
\small
\begin{tabular}{>{\raggedright\arraybackslash}p{0.165\textwidth}>{\raggedright\arraybackslash}p{0.27\textwidth}>{\raggedright\arraybackslash}p{0.23\textwidth}>{\raggedright\arraybackslash}p{0.23\textwidth}}
\toprule
\textbf{Boundary} & \textbf{User request} & \textbf{Boundary condition} & \textbf{PalmClaw response} \\
\midrule
Device permission & ``Add a meeting with Alex tomorrow at 10 a.m.'' & Required calendar permission is missing & Requests permission; executes only if granted \\
User confirmation & ``Take a photo of my desk to check the setup.'' & Photo capture requires user involvement & Opens the system camera for user completion \\
Workspace access & ``Save my shopping list to SDCard/list.txt.'' & Requested path requires all files access & Blocks the write; opens settings for manual access \\
Unsupported operation & ``Open Instagram to check updates from my friends.'' & No registered tool supports the GUI request & Reports that the operation is unsupported \\
\bottomrule
\end{tabular}
\end{table*}

\paragraph{Case 1: Calendar permission.}
In the calendar-permission case, the user asks PalmClaw to add a meeting, and the agent reaches the boundary when the calendar tool checks whether calendar access is available. As shown in Figure~\ref{fig:boundary-traces}, PalmClaw does not continue with an unchecked calendar operation. It sends a mobile permission request to the user, then continues the same task if access is granted. If access is not granted, the agent reports the permission issue and explains what the user needs to do instead of claiming that the meeting was created.

\paragraph{Case 2: Workspace boundary.}
In this case, the user asks PalmClaw to save a shopping list outside the current workspace, such as under \texttt{/sdcard/list.txt}. PalmClaw resolves the requested path before completing the file operation, and the file tool returns a boundary result because this public path requires all files access. Unlike calendar permission, this access cannot be granted through a normal mobile permission prompt. PalmClaw opens the application settings page and tells the user to enable all files access manually.

Together, these cases show three outcomes when a requested action reaches a boundary. PalmClaw can request a standard permission and resume, wait for the user to complete a required action, or stop and return a structured failure. In each case, the boundary is applied before the requested operation is executed, and the result is returned to the agent loop for its next decision.

\section{Conclusion}
\label{sec:conclusion}

In this paper, we examined how an agent framework can run natively on a mobile phone rather than treating the phone only as an interface to operate. We argue that mobile-agent design should consider both where the agent framework runs and how it accesses device capabilities. Guided by this view, we presented PalmClaw, an open-source native on-device agent framework that manages sessions, memory, skills, tools, and the agent loop directly on the phone. PalmClaw exposes device capabilities through device tools with explicit arguments, structured results, and tool-specific execution boundaries. Experiments show that PalmClaw improves task success by 11.5\% relative to the strongest baseline and reduces completion time by 94.9\%, while lowering setup burden. Trace-based cases further show how execution boundaries are applied before device operations are performed. We believe that treating the phone as an agent execution environment will broaden mobile-agent research from screen-control policies to complete on-device frameworks. This shift can support agents that use mobile capabilities directly while keeping their actions within explicit execution boundaries and under user control.

\section*{Limitations}

MobileTask focuses on tasks that can be completed through device-level capabilities and evaluated from the resulting state or answer. This scope supports clear outcome-based evaluation, but it does not cover third-party GUI workflows and produces a modest, unevenly distributed task set. The experiments also use one LLM backbone and one run per task; broader task categories, additional models, and repeated runs would provide a more complete measure of generality and robustness.

\section*{Ethics Statement}

PalmClaw is designed to keep mobile actions explicit and bounded: tools are registered, arguments are validated, sensitive actions rely on Android permissions or user confirmation, and file operations follow workspace boundaries. Deployments should still make provider-side data flow clear to users, especially when remote LLM providers receive prompts, tool results, or task context. The experiments in this paper use public or constructed evaluation tasks and do not require collecting private user data.

\section*{Acknowledgments}
The work described in this paper was supported by the Research Grants Council of Hong Kong (PolyU/15207122, PolyU/15209724, PolyU/15213323, PolyU/15205325), the PolyU internal grants (BDWP), and the PolyU–DIAGENS Joint Laboratory (ZG0Q).

Generative AI tools were used as assistants for software development and manuscript preparation. They supported code implementation, debugging, and review, as well as manuscript drafting, editing, and language polishing. All system designs, technical decisions, experiments, claims, and final content were determined, checked, and approved by the authors.


\bibliographystyle{unsrtnat} 
\bibliography{ref}

\clearpage
\appendix
\section{Additional Details}
\label{app:details}

\subsection{Datasets and Metrics}
\label{app:datasets}

\paragraph{MobileTask.}
MobileTask adapts tasks from AndroidWorld~\citep{rawles2024androidworlddynamicbenchmarkingenvironment}, MobileAgentBench~\citep{wang2024mobileagentbenchefficientuserfriendlybenchmark}, and Mobile-Bench~\citep{deng-etal-2024-mobile}. Each candidate is rewritten as a user-goal instruction with a task identifier, source label, expected artifact, and evaluation note. We retain a task only if 1) it can be attempted without a fixed GUI path, 2) it does not depend on third-party app state, and 3) it does not favor GUI-only agents. We also remove unstable cases, including relative dates, live weather, and duplicated calendar variants.

\paragraph{AssistantBench.}
AssistantBench~\citep{yoran2024assistantbenchwebagentssolve} is used as a secondary information-seeking set. From the 33-task dev set, we keep 19 tasks whose answers remain checkable after manual review. We exclude tasks with 1) unstable live-state answers, or 2) reference answers that cannot be checked reliably.

\paragraph{Metrics.}
MobileTask reports the mean success rate over 70 tasks. A task is successful when its final state or answer matches the predefined oracle, using deterministic checks for directly inspectable outcomes. For answer-based tasks, the same \texttt{DeepSeek-V4-Flash} model serves as the judge. It receives the task instruction, the agent's final answer, and fixed reference evidence, and returns a binary success-or-failure decision. AssistantBench reports official accuracy averaged over the subset, using word-level F1 for string answers, log-distance scoring for numeric answers, and aligned mean F1 for list answers.

\begin{figure}[t]
\centering
\includegraphics[width=1\linewidth]{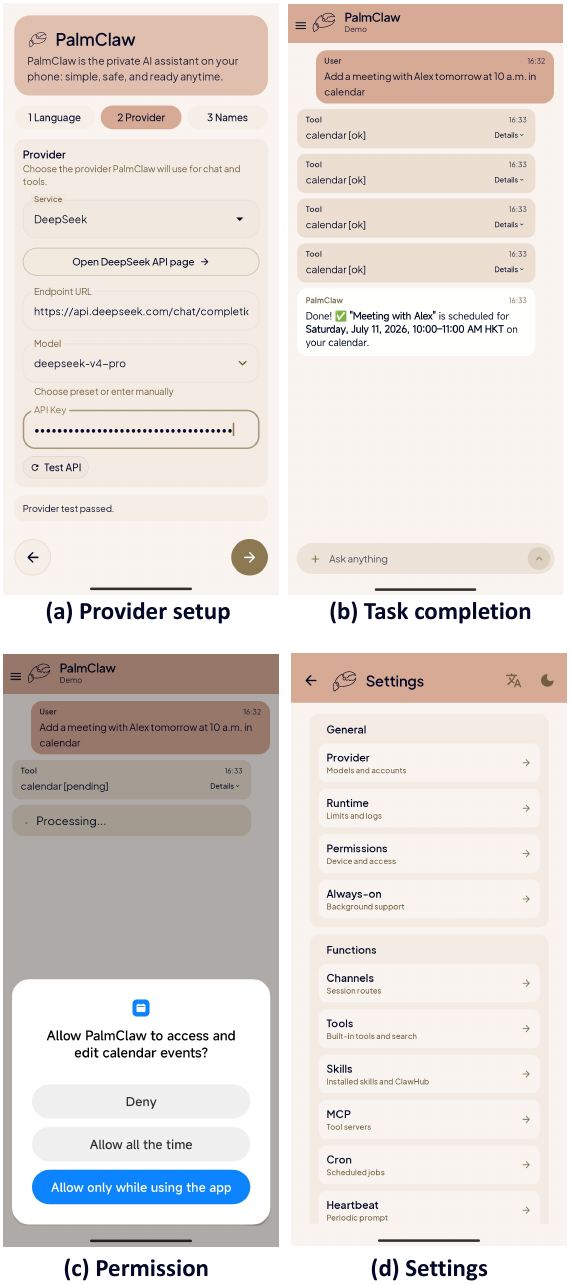}
\caption{PalmClaw application interface. The screenshots show (a) provider setup, (b) a calendar task completed through device tools, (c) a permission request before calendar access, and (d) settings.}
\label{fig:app-screenshots}
\end{figure}

\end{document}